\documentclass{article}
\usepackage{ijcai17}

\usepackage{times}
\usepackage{latexsym}

\usepackage{url}

\usepackage{amsmath}
\usepackage{amssymb}
\usepackage{amsopn}
\usepackage{subcaption}
\usepackage{CJKutf8}
\usepackage{graphicx}
\usepackage{multirow}
\usepackage{color}
\usepackage{tabularx}
\usepackage{enumitem}
\usepackage{algorithm}  
\usepackage{algorithmic}  
\usepackage[skip=2pt,belowskip=-12pt,aboveskip=2pt,font=small]{caption}

\newcommand{\chinese}[1]{{\begin{CJK*}{UTF8}{gkai} #1 \end{CJK*}}}
\definecolor{zb_red}{RGB}{200, 0, 0}

\usepackage{enumitem}
\setenumerate[1]{itemsep=0.30pt,partopsep=0pt,parsep=\parskip,topsep=5pt}
\setitemize[1]{itemsep=0.30pt,partopsep=0pt,parsep=\parskip,topsep=5pt}
\setdescription{itemsep=0.30pt,partopsep=0pt,parsep=\parskip,topsep=5pt} 

\title{A GRU-Gated Attention Model for Neural Machine Translation\thanks{\textsuperscript{\textcopyright} 20xx IEEE. Personal use of this material is permitted. Permission
from IEEE must be obtained for all other uses, in any current or future
media, including reprinting/republishing this material for advertising or
promotional purposes, creating new collective works, for resale or
redistribution to servers or lists, or reuse of any copyrighted
component of this work in other works.}}
\author{Biao Zhang$^{1}$, Deyi Xiong$^{2}$ and Jinsong Su$^{1}$\\
	Xiamen University, Xiamen, China 361005$^{1}$ \\
	Soochow University, Suzhou, China 215006$^{2}$ \\
	{\tt zb@stu.xmu.edu.cn, jssu@xmu.edu.cn} \\
	{\tt dyxiong@suda.edu.cn} \\
}

\begin{document}

\maketitle

\begin{abstract}
Neural machine translation (NMT) heavily relies on an attention network to produce a context vector for each target word prediction. In practice, we find that context vectors for different target words are quite similar to one another and therefore are insufficient in discriminatively predicting target words. The reason for this might be that context vectors produced by the vanilla attention network are just a weighted sum of source representations that are invariant to decoder states. In this paper, we propose a novel GRU-gated attention model (GAtt) for NMT which enhances the degree of discrimination of context vectors by enabling source representations to be sensitive to the partial translation generated by the decoder. GAtt uses a gated recurrent unit (GRU) to combine two types of information: treating a source annotation vector originally produced by the bidirectional encoder as the history state while the corresponding previous decoder state as the input to the GRU. The GRU-combined information forms a new source annotation vector. In this way, we can obtain translation-sensitive source representations which are then feed into the attention network to generate discriminative context vectors. We further propose a variant that regards a source annotation vector as the current input while the previous decoder state as the history. Experiments on NIST Chinese-English translation tasks show that both GAtt-based models achieve significant improvements over the vanilla attention-based NMT. Further analyses on attention weights and context vectors demonstrate the effectiveness of GAtt in improving the discrimination power of representations and handling the challenging issue of over-translation.
\end{abstract}

\section{Introduction}

Neural machine translation (NMT), as a large, single and end-to-end trainable neural network, has attracted wide attention in recent years~\cite{DBLP:journals/corr/SutskeverVL14,DBLP:journals/corr/BahdanauCB14,DBLP:journals/corr/ShenCHHWSL15,jean-EtAl:2015:ACL-IJCNLP,luong-EtAl:2015:ACL-IJCNLP,DBLP:journals/corr/WangLTLXZ16}. Currently, most NMT systems use an encoder to read a source sentence into a vector and a decoder to map the vector into the corresponding target sentence. What makes NMT outperform conventional statistical machine translation (SMT) is the attention mechanism~\cite{DBLP:journals/corr/BahdanauCB14}, an information bridge between the encoder and the decoder that produces context vectors by dynamically detecting relevant source words for predicting the next target word.

\begin{figure}[t]
\centering
\includegraphics[scale=0.41]{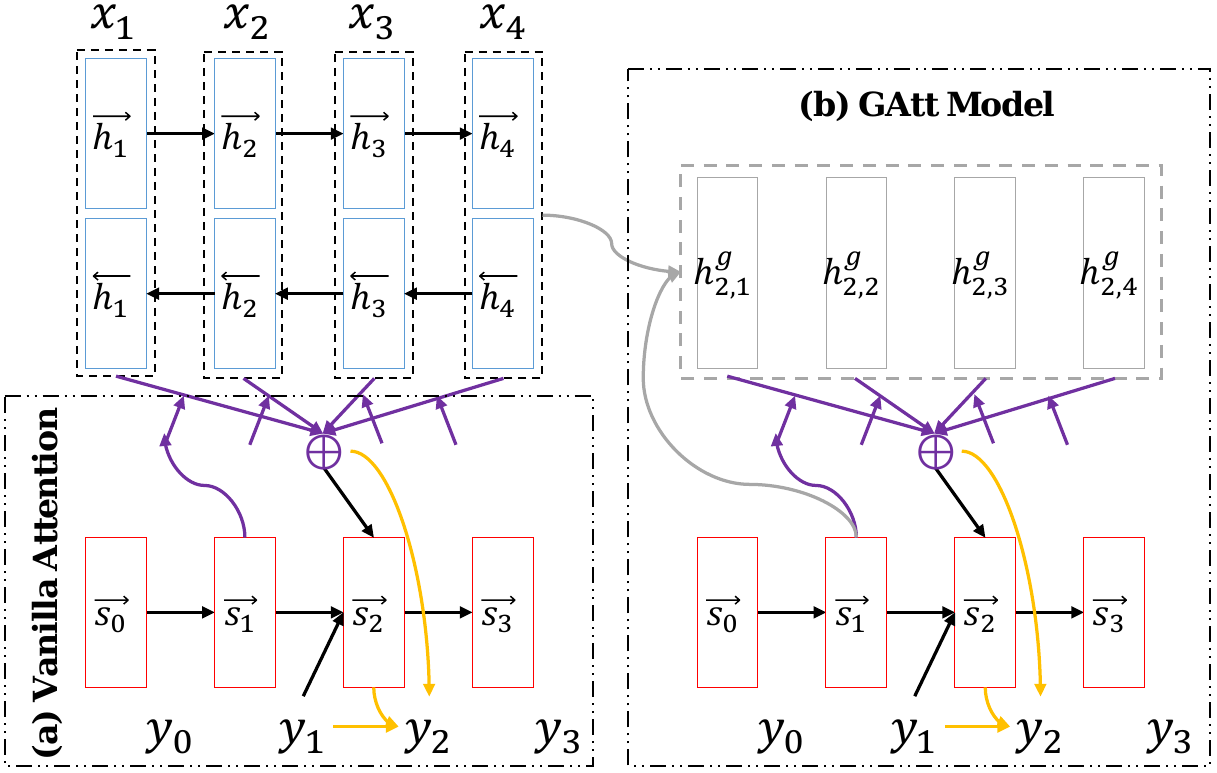}
\caption{\label{overall} Illustration of the vanilla attention and the proposed GAtt. We use blue and red color to indicate the source and target side respectively. The yellow and purple color denotes the information flow for target word prediction and attention respectively, and the gray color denotes GRU-gated layer. The doted boxes show the vanilla-attention-specific and GAtt-specific components.}
\end{figure}
Intuitively, different target words would be aligned to different source words so that the generated context vectors differ significantly from one another across different decoding steps. In other words, these context vectors should be discriminative enough for target word prediction otherwise the same target words might be generated repeatedly (a well-known issue of NMT: over-translation, see Section \ref{over_translation_analysis_sec}). However, this is often not true in practice, even when ``attended'' source words are rather relevant. We observe that (see Section \ref{trans_analysis_sec}) the context vectors are very similar to each other, and that the variance in each dimension of these vectors across different decoding steps is very small. These indicate that the vanilla attention mechanism suffers from its inadequacy in distinguishing different translation predictions. The reason behind, we conjecture, lies in the architecture of the attention mechanism which simply calculates a linearly weighted sum of source representations that are invariant across decoding steps. Such invariance in source representations may lead to the undesirable small variance of context vectors.

In order to handle this issue, in this paper, we propose a novel GRU-gated attention model (GAtt) for NMT. The key is that we can increase the degree of variance in context vectors by refining source representations according to the partial translation generated by the decoder. The refined source representations are composed of the original source representations and the previous decoder state at each decoding step. We show the overall framework of our model and highlight the difference between GAtt and the vanilla attention in Figure \ref{overall}. GAtt significantly extends the vanilla attention by inserting a gating layer between the encoder and the vanilla attention network. Specifically, we model this gating layer with a GRU unit~\cite{journals/corr/ChungGCB14}, which takes the original source representations as its history and the corresponding previous decoder state as its current input. In this way, GAtt can produce translation-sensitive source representations so as to improve the variance in context vectors and therefore its discrimination ability in target word prediction.

As GRU is able to control the information flow between the history and current input through its reset and update gate, we further propose a variant of GAtt that, instead, regards the previous decoder state as the history while the original source representations as the current inputs. Both models are simple yet efficient in training and decoding. 

We testify GAtt on Chinese-English translation tasks. Experimental results show that both GAtt-based models significantly outperform the vanilla attention-based NMT. We further analyze the generated attention weights and context vectors, showing that the attention weights are more accurate and the context vectors are more discriminative for target word prediction.

\section{Related Work}

Our work contributes to the development of attention mechanism in NMT. Originally, NMT does not have the attention mechanism and mainly relies on the encoder to summarize all source-side semantic details into a fixed-length vector~\cite{DBLP:journals/corr/SutskeverVL14,cho-EtAl:2014:EMNLP2014}. Bahdanau et al.~\shortcite{DBLP:journals/corr/BahdanauCB14} find that using a fixed-length vector, however, is not adequate to represent a source sentence and propose the popular attention mechanism, enabling the model to automatically search for parts of a source sentence that are relevant to the next target word. From then on, the attention mechanism has gained extensive concern. Luong et al.~\shortcite{luong-pham-manning:2015:EMNLP} explore several effective approaches to the attention network, introducing the local and global attention model. Tu et al.~\shortcite{DBLP:journals/corr/TuLLLL16} introduce a coverage vector to keep track of the attention history such that the attention network can pay more attention to untranslated source words. Mi et al.~\shortcite{mi-wang-ittycheriah:2016:EMNLP2016} leverage well-trained word alignments to directly supervise the attention weights in NMT. Yang et al.~\shortcite{2016arXiv160705108Y} bring a recurrence along the context vector to help adjust the future attention. Cohn et al.~\shortcite{DBLP:journals/corr/CohnHVYDH16} incorporate several structural bias, such as position bias, markov condition and fertilities, into the attention-based neural translation model. However, all these models mainly focus on how to make the attention weights more accurate. As we mentioned, even with well-designed attention models, context vectors may be lack of discrimination ability for target word prediction.

Another closely related work is the interactive attention model~\cite{meng-EtAl:2016:COLING} which treats source representations as a memory and models the interaction between the decoder and this memory during translation via reading and writing operations. To some extent, our model can also be regarded as a memory network, which only includes the reading operation. However, our reading operation differs significantly from that in the interactive attention, where we employ the GRU unit for composition while they merely use the content-based addressing. Compared with the interactive attention, our GAtt, without the writing operation, is more efficient in both training and decoding.

The gate mechanism in our GAtt is built on the GRU unit. GRU usually acts as a recurrent unit that leverages a reset gate and an update gate to control how much information flow from the history state and the current input respectively~\cite{journals/corr/ChungGCB14}. It is an extension of the vanilla recurrent neural network (RNN) unit with the advantage of alleviating the vanishing and exploding gradient problems during training~\cite{Bengio-trnn94}, and also a simplification of the LSTM model~\cite{Hochreiter:1997:LSM:1246443.1246450} with the advantage of efficient computation. The idea of using GRU as a gate mechanism, to the best of our knowledge, has never been investigated before. 

Additionally, our model is also related with the tree-structured LSTM~\cite{tai-socher-manning:2015:ACL-IJCNLP}, where LSTM is adapted to compose vary-sized children nodes and current input node in a dependency tree into the current hidden state. GAtt differs significantly from the tree-structured LSTM in that the latter employs the sum operation to deal with the vary-sized representations, while our model leverages the attention mechanism. 

\section{Background}

In this section, we briefly review the vanilla attention-based NMT~\cite{DBLP:journals/corr/BahdanauCB14}. Unlike conventional SMT, NMT directly maps a source sentence $\mathbf{x} = \{x_1, \ldots, x_n\}$ to its target translation $\mathbf{y} = \{y_1, \ldots, y_m\}$ using an encoder-decoder framework. The encoder reads the source sentence $\mathbf{x}$, and encodes the representation of each word $\mathbf{h}_i$ by summarizing the information of neighboring words. As shown by the blue color in Figure \ref{overall}, this is achieved by a bidirectional RNN, specifically the bidirectional GRU model. 

The decoder is a conditional language model which generates the target sentence word by word using the following conditional probability (see the yellow lines in Figure \ref{overall} (a)):
\begin{equation}\label{conditional_prediction}
p(y_j|\mathbf{x}, \mathbf{y}_{<j})=softmax\left(g(E_{y_{j-1}}, \mathbf{s}_{j}, \mathbf{c}_{j})\right)
\end{equation}
where $\mathbf{y}_{<j} = \{y_1, \cdots, y_{j-1}\}$ is a partial translation, $E_{y_{j-1}} \in \mathbb{R}^{d_w}$ is the embedding of previously generated target word $y_{j-1}$, $\mathbf{s}_{j} \in \mathbb{R}^{d_h}$ is the $j$-th target-side decoder state and $g(\cdot)$ is a highly non-linear function. Please refer to \cite{DBLP:journals/corr/BahdanauCB14} for more details. What we concern in this paper is $\mathbf{c}_j \in \mathbb{R}^{2d_h}$, which is the translation-sensitive context vector produced by the attention mechanism.

\noindent {\bf Attention Mechanism} acts as a bridge between the encoder and the decoder, which makes them tightly coupled. The attention network aims at recognizing which source words are relevant to the next target word and giving high attention weights to these words in computing the context vector $\mathbf{c}_j$. This is based on the encoded source representations $\mathbf{H} = \{\mathbf{h}_1, \cdots, \mathbf{h}_n\}$ and the previous decoder state $\mathbf{s}_{j-1}$ (see the purple color in Figure \ref{overall} (a)). Formally,
\begin{equation}\label{original_att}
\mathbf{c}_j = \text{Att}(\mathbf{H}, \mathbf{s}_{j-1})
\end{equation}
$\text{Att}(\cdot)$ denotes the whole process. It first computes an attention weight $\alpha_{ji}$ to measure the degree of relevance of a source word $x_i$ for predicting the target word $y_j$ via a feed-forward neural network:
\begin{equation}
\alpha_{ji} = \frac{\exp\left(e_{ji}\right)}{\sum_k \exp\left(e_{jk})\right)}
\end{equation}
The relevance score $e_{ji}$ is estimated via an {\em alignment model} as in~\cite{DBLP:journals/corr/BahdanauCB14}: $e_{ji} = v^T_a \tanh(W_a \mathbf{s}_{j-1} + U_a \mathbf{h}_i)$. Intuitively, the higher attention weight $\alpha_{ji}$ is, the more important word $x_i$ is for the next word prediction. Therefore, $\text{Att}(\cdot)$ generates $\mathbf{c}_j$ by directly weighting the source representations $\mathbf{H}$ with their corresponding attention weights $\{\alpha_{ji}\}_{i=1}^{n}$:
\begin{equation}\label{context_vector}
\mathbf{c}_j = \sum_i \alpha_{ji} \mathbf{h}_i
\end{equation}

Although this vanilla attention model is very successful, we find that, in practice, the resulted context vectors $\{\mathbf{c}_j\}_{j=1}^{m}$ are very similar to one another. In other words, these context vectors are not discriminative enough. This is undesirable because it makes the decoder (Eq. (\ref{conditional_prediction})) hesitate in deciding which target word should be predicted. We attempt to solve this problem in the next section. 

\section{GRU-Gated Attention for NMT}

The problem mentioned above reveals some shortcomings of the vanilla attention mechanism. Let's revisit the generation of $\mathbf{c}_j$ in Eq. (\ref{context_vector}). As different target words might be aligned to different source words, the attention weights of source words vary across different decoding steps. However, no matter how the attention weights of source words vary, the source representations $\mathbf{H}$ remain the same, i.e. they are decoding-invariant. And this invariance would limit the discrimination power of the generated context vectors. 

Accordingly, we attempt to break up this invariance by refining the source representations before they are input to the vanilla attention  network at each decoding step. To this end, we propose the GRU-gated attention (GAtt), which, similar to the vanilla attention, can be formulated into the following form:
\begin{equation}
\mathbf{c}_j = \text{GAtt}(\mathbf{H}, \mathbf{s}_{j-1})
\end{equation}
The gray color in Figure \ref{overall} (b) highlights the major difference between GAtt and the vanilla attention. Specifically, GAtt consists of two layers: a {\em gating layer} and an {\em attention layer}.

{\bf Gating Layer.} This layer aims at refining the source representations according to the previous decoder state $\mathbf{s}_{j-1}$ so as to compute translation-relevant source representations. Formally,
\begin{equation}
\mathbf{H}^g_j = \text{Gate}(\mathbf{H}, \mathbf{s}_{j-1})
\end{equation}
The $\text{Gate}(\cdot)$ should be capable of dealing with the complex interactions between the source sentence and the partial translation, and freely controlling the semantic match and information flow between them. Instead of using conventional gating mechanism~\cite{journals/corr/ChungGCB14}, we directly choose the whole GRU unit to perform this task. For a source representation $\mathbf{h}_i$, GRU treats it as the history representation and refines it using the current input, i.e. the previous decoder state $\mathbf{s}_{j-1}$:
\begin{equation} \label{gru_model}
\begin{split}
\mathbf{z}_{ji} = \sigma(W_z \mathbf{s}_{j-1} + U_z \mathbf{h}_{i} + b_z) \\
\mathbf{r}_{ji} = \sigma(W_r \mathbf{s}_{j-1} + U_r \mathbf{h}_{i} + b_r) \\
\overline{h}_{ji} = \tanh(W \mathbf{s}_{j-1} + U \left[ \mathbf{r}_{ji} \odot \mathbf{h}_{i} \right] + b) \\
\mathbf{h}^g_{ji} = (1 - \mathbf{z}_{ji}) \odot \mathbf{h}_{i} + \mathbf{z}_{ji} \odot \overline{h}_{ji}
\end{split}
\end{equation}
where $\sigma(\cdot)$ is the sigmoid function, and $\odot$ denotes the element-wise multiplication. Intuitively, the reset gate $\mathbf{r}_{ji}$ and update gate $\mathbf{z}_{ji}$ measure the degree of the semantic match between the source sentence and partial translation. The former determines how much the original source information could be used to combine the partial translation, while the latter defines how much the original source information can be kept around. As a result, $\mathbf{h}^g_{ji}$ becomes translation-sensitive, rather than decoding-invariant, which is desired to strengthen the discrimination power of $\mathbf{c}_j$.

\begin{table*}[t]
\begin{center}
{\small
\begin{tabular}{c|c||c|lllllll}
\multicolumn{1}{c|}{\bf Metric} &
\multicolumn{1}{c||}{\bf System} &
\multicolumn{1}{c|}{\bf MT05 } &
\multicolumn{1}{l}{\bf MT02 } &
\multicolumn{1}{l}{\bf MT03 } &
\multicolumn{1}{l}{\bf MT04 } &
\multicolumn{1}{l}{\bf MT06 } &
\multicolumn{1}{l}{\bf MT08 } &
\multicolumn{1}{l}{\bf AVG} \\
\hline
\hline
\multirow{4}{*}{\bf BLEU} 
	& {\it Moses} & 31.70 & 33.61 & 32.63 & 34.36 & 31.00 & 23.96 & 31.11 \\

	& {\it RNNSearch} & 34.72 & 37.95 & 35.23 & 37.32 & 33.56 & 26.12 & 34.04 \\
\cline{2-9}

	& {\it GAtt} & 36.20 & {\bf 39.30}$^{\scriptsize \Uparrow+}$ & {\bf 37.22}$^{\scriptsize \Uparrow++}$ & 39.61$^{\scriptsize \Uparrow++}$ & 34.81$^{\scriptsize \Uparrow+}$ & {\bf 27.56}$^{\scriptsize \Uparrow+}$ & 35.70 \\

	& {\it GAtt-Inv} & 36.50 & {38.76}$^{\scriptsize \Uparrow+}$ & {36.89}$^{\scriptsize \Uparrow++}$ & {\bf 40.09}$^{\scriptsize \Uparrow++}$ & {\bf 35.46}$^{\scriptsize \Uparrow++}$ & 27.32$^{\scriptsize \Uparrow+}$ & {\bf 35.70} \\
\hline
\hline
\multirow{4}{*}{\bf TER}
	& {\it Moses} & 58.24 & 57.81 & 57.22 & 56.11 & 56.90 & 60.33 & 57.67 \\

	& {\it RNNSearch} & 59.58 & 56.48 & 58.47 & 56.55 & 57.72 & 61.68 & 58.18 \\
\cline{2-9}

	& {\it GAtt} & 57.25 & {\bf 54.32} & 56.13 & 54.15 & 55.90 & {\bf 59.82} & 56.06 \\

	& {\it GAtt-Inv} & 56.67 & { 54.66} & {\bf 56.05} & {\bf 53.69} & {\bf 55.64} & {59.93} & {\bf 55.99} \\
\end{tabular}
}
\end{center}
\caption{\label{test_performance} BLEU and TER scores on the Chinese-English translation tasks. {\bf AVG} = average BLEU/TER score on all test sets. We highlight the best results in bold for each test set. ``$\uparrow$/$\Uparrow$'': significantly better than {\it Moses} ($p$ $<$ $0.05$/$p$ $<$ $0.01$); ``$+$/$++$'': significantly better than {\it GroundHog} ($p$ $<$ $0.05$/$p$ $<$ $0.01$). Higher BLEU scores and lower TER scores indicate better translation quality.}
\end{table*}
\begin{table*}[t]
\begin{center}
{\small
\begin{tabular}{c|c||c|lllllll}
\multicolumn{1}{c|}{\bf Metric} &
\multicolumn{1}{c||}{\bf System} &
\multicolumn{1}{c|}{\bf MT05 } &
\multicolumn{1}{l}{\bf MT02 } &
\multicolumn{1}{l}{\bf MT03 } &
\multicolumn{1}{l}{\bf MT04 } &
\multicolumn{1}{l}{\bf MT06 } &
\multicolumn{1}{l}{\bf MT08 } &
\multicolumn{1}{l}{\bf AVG} \\
\hline
\hline
\multirow{4}{*}{\bf BLEU} 
	& {\it RNNSearch+GAtt} & 38.13 & 41.13 & 38.95 & 41.17 & 36.75 & 29.28 & 37.46 \\

	& {\it RNNSearch+GAtt-Inv} & 37.54 & 40.67 & 38.84 & 41.12 & 37.10 & 28.94 & 37.33 \\
	
	& {\it GAtt+GAtt-Inv} & 38.88 & 40.78 & 38.72 & 42.08 & 37.54 & 29.59 & 37.94 \\

	& {\it RNNSearch+GAtt+GAtt-Inv} & {\bf 39.24} & {\bf 41.97} & {\bf 40.01} & {\bf 42.67} & {\bf 38.19} & {\bf 30.34} & {\bf 38.64} \\
\hline
\hline
\multirow{4}{*}{\bf TER}
	& {\it RNNSearch+GAtt} & 56.05 & 53.68 & 55.03 & 53.30 & 54.71 & 58.98 & 55.14 \\

	& {\it RNNSearch+GAtt-Inv} & 55.95 & 53.30 & 55.08 & 53.27 & 54.86 & 59.15 & 55.13 \\
	
	& {\it GAtt+GAtt-Inv} & 55.04 & {52.82} & {\bf 54.20} & {\bf 52.24} & 53.99 & 58.27 & 54.30 \\

	& {\it RNNSearch+GAtt+GAtt-Inv} & {\bf 54.74} & {\bf 52.59} & 54.22 & {52.32} & {\bf 53.76} & {\bf 57.88} & {\bf 54.15} \\
\end{tabular}
}
\end{center}
\caption{\label{test_ensemble} BLEU and TER scores of the ensemble of different NMT systems.}
\end{table*}
{\bf Attention Layer.} This layer is the same as the vanilla attention mechanism:
\begin{equation}\label{GAtt_att}
\mathbf{c}_j = \text{Att}(\mathbf{H}^g_{j}, \mathbf{s}_{j-1})
\end{equation}
The $\text{Att}(\cdot)$ in Eq. (\ref{GAtt_att}) denotes the same procedure as that in Eq. (\ref{original_att}). However, instead of paying attention to the original source representations $\mathbf{H}$, this layer relies on the gate-refined source representations $\mathbf{H}^g_{j}$. Notice that $\mathbf{H}^g_j$ is adaptive during decoding, indicated with the subscript $j$. Ideally, we expect $\mathbf{H}^g_{j}$ is decoding-specific enough such that $\mathbf{c}_j$ can vary significantly across different target words.

Notice that $\text{Gate}(\cdot)$ is not a multi-stepped RNN. It is simply a composition function, or only one-stepped RNN. Therefore, it is computationally efficient. To train our model, we employ the standard training objective, i.e. maximizing the log-likelihood of the training data, and optimize the model parameters using the standard stochastic gradient algorithm.    

{\bf Model Variant} We refer to the above model as {\em GAtt}, which regards the source representations as the history and the previous decoder state as the current input. Which information should be treated as input or history does not matter, especially for the GRU unit since GRU is able to control the information flow freely. We can also use the previous decoder state as the history and the source representations as the current input. We refer to this model as {\em GAtt-Inv}. Formally,
\begin{equation}
\mathbf{c}_j = \text{GAtt-Inv}(\mathbf{H}, \mathbf{s}_{j-1})
\end{equation}
with,
\begin{equation*}
\mathbf{c}_j = \text{Att}({\mathbf{H}^g_{j}}^{\prime}, \mathbf{s}_{j-1})\quad{\mathbf{H}^g_{j}}^{\prime} = \text{Gate}(\mathbf{s}_{j-1}, \mathbf{H})
\end{equation*}
The major difference lies at the order of the inputs in $\text{Gate}(\cdot)$, since the inputs to GRU are directional. We verify both model variants through the following experiments.

\section{Experiments}

\begin{table*}[t]
\begin{center}
{\small
\begin{tabular}{c||p{1.35\columnwidth}}
\hline
\hline
{\it Source} & \chinese{他~说~,~难民~\textcolor{zb_red}{重新~融入~社会}~是~\textcolor{zb_red}{临时}~政府~的~工作~重点~之一~。} \\
\hline
\multirow{2}{*}{\it Reference} & {\it he said the refugees ' \textcolor{zb_red}{re-integration into society} is one of the top priorities on the \textcolor{zb_red}{interim} government 's agenda .} \\
\hline
{\it RNNSearch} & {\it he said that refugee is one of the key government tasks .}\\
\hline
\multirow{2}{*}{\it GAtt} & {\it he said that the \textcolor{zb_red}{re - integration of the refugees} was one of the key tasks of the \textcolor{zb_red}{interim} government .} \\
\hline
\multirow{2}{*}{\it GAtt-Inv} & {\it he said that the refugees ' \textcolor{zb_red}{integration into society} is one of the key tasks of the \textcolor{zb_red}{interim} government .} \\
\hline
\hline
\end{tabular}
}
\end{center}
\caption{\label{sentence_analysis} Example translations generated by different systems. We use red color to highlight interesting parts.}
\end{table*}
\begin{figure*}[t]
	\captionsetup[subfigure]{skip=12pt,belowskip=3pt,aboveskip=6pt}
   	\centering
  	\begin{subfigure}{.35\textwidth}
  		\centering
  		\includegraphics[width=\textwidth]{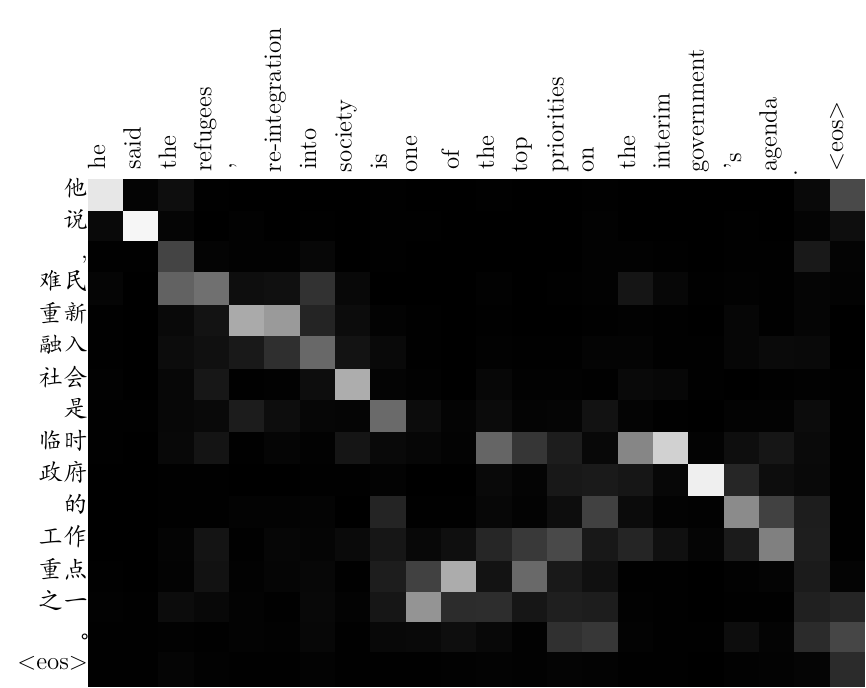}
  	\end{subfigure}\quad
  	\begin{subfigure}{.35\textwidth}
  		\centering
  		\includegraphics[width=\textwidth]{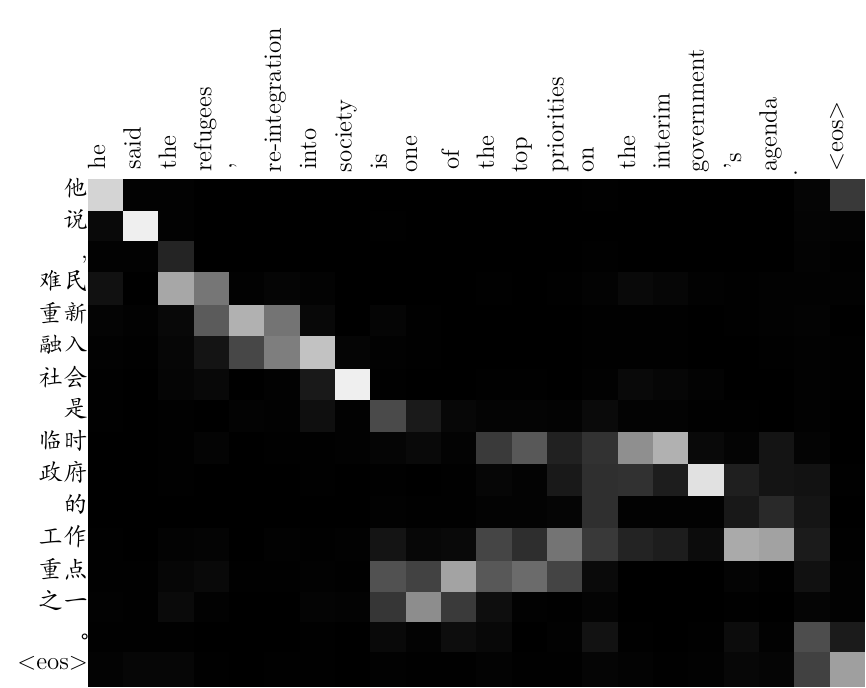}
  	\end{subfigure}
  	\caption{\label{visual_work} Visualization of the attention weights for RNNSearch (left) and GAtt (right). }
\end{figure*}
\begin{figure*}[t]
	\captionsetup[subfigure]{skip=12pt,belowskip=3pt,aboveskip=6pt}
	\centering
	\begin{subfigure}{0.38\textwidth}
		\centering
		\includegraphics[width=\textwidth]{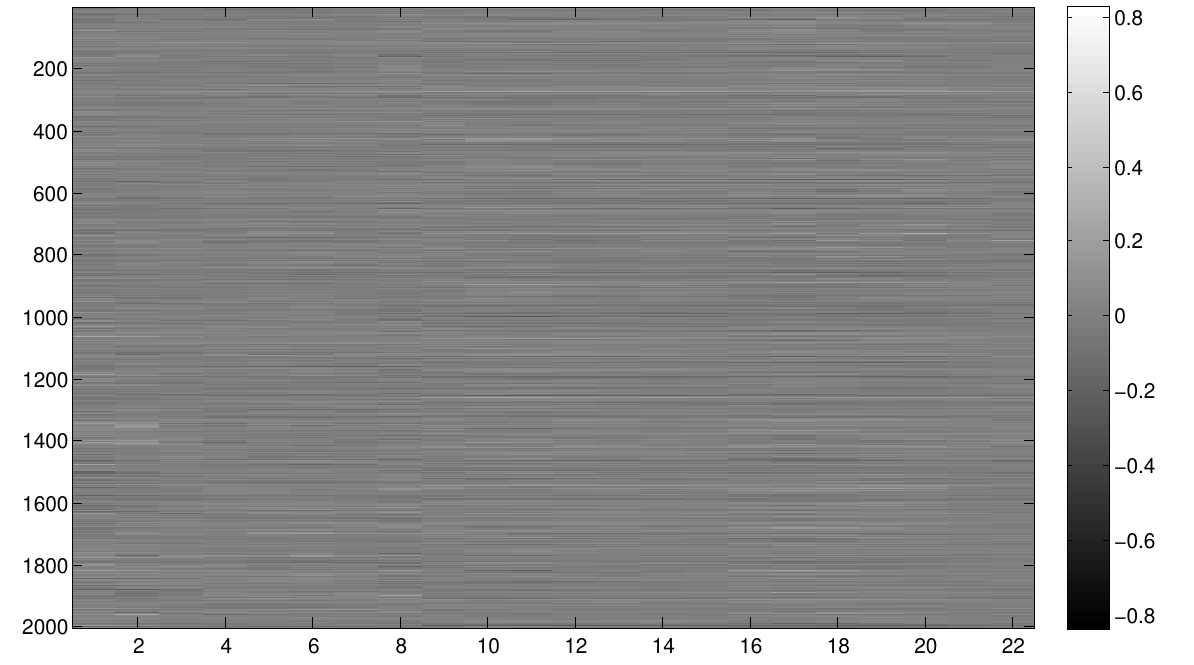}
	\end{subfigure}\quad
	\begin{subfigure}{0.38\textwidth}
		\centering
		\includegraphics[width=\textwidth]{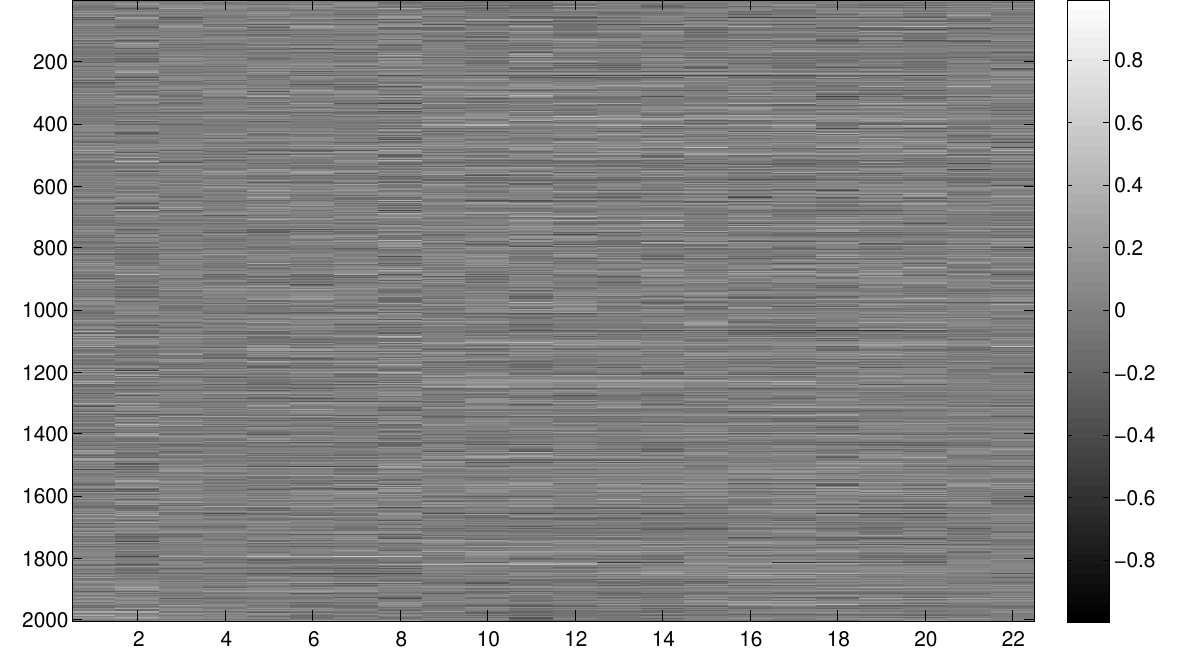}
	\end{subfigure}
	\caption{\label{context_visual} Visualization of context vectors for RNNSearch (left) and GAtt (right) generated as in Figure \ref{visual_work}. The horizontal and vertical axis denotes the translation and context dimension respectively.} 
\end{figure*}
\subsection{Setup}

We evaluated the effectiveness of our model on Chinese-English translation tasks. Our training data consists of 1.25M sentence pairs, with 27.9M Chinese words and 34.5M English words respectively\footnote{This data is a combination of LDC2002E18, LDC2003E07, LDC2003E14, Hansards portion of LDC2004T07, LDC2004T08 and LDC2005T06.}. We chose the NIST 2005 dataset as the development set to perform model selection, and the NIST 2002, 2003, 2004, 2006 and 2008 datasets as our test sets. There are 878, 919, 1788, 1082 and 1664 sentences in NIST 2002, 2003, 2004, 2005, 2006, 2008 dataset respectively. We evaluated the translation quality using the case-insensitive BLEU-4 metric~\cite{PapineniEtAl2002}\footnote{https://github.com/moses-smt/mosesdecoder/blob/master/scripts/\\generic/multi-bleu.perl} and TER metric~\cite{Snover06astudy}\footnote{http://www.cs.umd.edu/~snover/tercom/}. We performed paired bootstrap sampling~\cite{koehn04} for statistical significance test using the script in {\em Moses}\footnote{https://github.com/moses-smt/mosesdecoder/blob/master/scripts/\\analysis/bootstrap-hypothesis-difference-significance.pl}.

\subsection{Baselines}

We compared our proposed model against the following two state-of-the-art SMT and NMT systems:
\begin{itemize}
\item
{\it Moses}~\cite{Koehn:2007:MOS:1557769.1557821}: an open source state-of-the-art phrase-based SMT system.
\item
{\it RNNSearch}~\cite{DBLP:journals/corr/BahdanauCB14}: a state-of-the-art attention-based NMT system using the vanilla attention mechanism. We further feed the information of $y_{j-1}$ to the attention, and implemented the decoder with two GRU layers, following the suggestions in {\em dl4mt}\footnote{https://github.com/nyu-dl/dl4mt-tutorial/tree/master/session3}.
\end{itemize}

For {\it Moses}, we trained a 4-gram language model on the target portion of training data using the SRILM\footnote{http://www.speech.sri.com/projects/srilm/download.html} toolkit with modified Kneser-Ney smoothing. The word alignments were obtained with GIZA++~\cite{Och:2003:SCV:778822.778824} on the training corpora in both directions, using the ``grow-diag-final-and'' strategy~\cite{Koehn:2003:SPT:1073445.1073462}. All other parameters were kept as the default settings.

For {\it RNNSearch}, we limit the vocabulary of both source and target languages to be the most frequent 30K words, covering approximately 97.7\% and 99.3\% of the two corpora respectively. The words that do not appear in the vocabulary were mapped to a special token ``{\it UNK}''. We trained our model with the sentences of length up to 50 words in the training data. Following the settings in~\cite{DBLP:journals/corr/BahdanauCB14}, we set $d_w$ $=$ $620$, $d_h$ $=$ $1000$. We initialized all parameters randomly according to a normal distribution ($\mu=0, \sigma=0.01$) except the square matrices which are initialized with random orthogonal matrices. We used the Adadelta algorithm~\cite{DBLP:journals/corr/abs-1212-5701} for optimization, with a batch size of $80$ and gradient norm as $5$. The model parameters were selected according to the maximum BLEU points on the development set. Additionally, during decoding, we used the beam-search algorithm, and set the beam size to 10.

For {\it GAtt}, we randomly initialized its parameters as what we do in {\it RNNSearch}. All the other settings are the same as {\it RNNSearch}. All NMT systems were trained on a GeForce GTX 1080 using the computational framework {\em Theano}. In one hour, the {\it RNNSearch} system processes about 2769 batches while {\it GAtt} processes 1549 batches.

\subsection{Translation Results}

The results are summarized in Table \ref{test_performance}. Both GAtt and GAtt-Inv outperform both Moses and RNNSearch. Specially, GAtt yields 35.70 BLEU and 56.06 TER scores on average, with improvements of 4.59 BLEU and 1.61 TER points over Moses, and 1.66 BLEU and 2.12 TER points over RNNSearch; GAtt-Inv achieves 35.70 BLEU and 55.99 TER scores on average, with gains of 4.59 BLEU and 1.68 TER points over Moses, and 1.66 BLEU and 2.19 TER points over RNNSearch. All improvements are statistically significant. 

It seems that GAtt-Inv obtains very slightly better performance than GAtt in terms of TER on average. However, these improvements are neither significant nor consistent. In other words, GAtt is as efficient as GAtt-Inv. This is reasonable, since the difference of GAtt and GAtt-Inv lies at the order of inputs to GRU, and GRU is able to control its information flow from each input through its reset and update gate. 

\subsection{Effects of Model Ensemble}

We further testify whether the ensemble of our models and RNNSearch can generate better performance against any single system. We ensemble different systems by simply averaging their predicted target word probabilities at each decoding step, as suggested in~\cite{luong-EtAl:2015:ACL-IJCNLP}. We show the results in Table \ref{test_ensemble}. Not surprisingly, all the ensemble systems achieves significant improvements over the best single system. And the ensemble of ``{\em RNNSearch+GAtt+GAtt-Inv}'' produces the best results, 38.64 BLEU and 54.15 TER scores on average. This demonstrates that these neural models are complementary and beneficial to each other.

\subsection{Translation Analysis}\label{trans_analysis_sec}

In order to have a deep understanding of how the proposed models work, we dug into the translated sentences of different neural systems. Table \ref{sentence_analysis} shows an example. All the neural models generate very fluent translations. However, RNNSearch only translates the rough meaning of the source sentence, ignoring important sub-phrases ``\chinese{重新~融入~社会}'' and ``\chinese{临时}''. These missing translations resonate with the finding of Tu et al.~\shortcite{DBLP:journals/corr/TuLLLL16}. sIn contrast, GAtt and GAtt-Inv are able to capture these two sub-phrases, generating the key translations ``{\em integration}'' and ``{\em interim}''. 

\begin{table}[t]
\begin{center}
{\small
\begin{tabular}{c||c|c}
{\bf System} & {\bf SAER} & {\bf AER} \\
\hline
\hline
{\em Tu et al.~\shortcite{DBLP:journals/corr/TuLLLL16}} & 64.25 & 50.50 \\
{\em RNNSearch} & 64.10 & 50.83 \\
\hline
{\em GAtt} & {\bf 56.19} & {\bf 43.53} \\
{\em GAtt-Inv} & { 58.29} & { 43.60} \\
\end{tabular}
}
\end{center}
\caption{\label{align_quality} SAER and AER scores of word alignments deduced by different neural systems. The lower the score, the better the alignment quality.}
\end{table}
To find the underlying reason, we investigated their generated attention weights. Rather than using the generated target sentences, we feed the same reference translations into RNNSearch and GAtt for making a fair comparison\footnote{We do not analyze GAtt-Inv because it is very similar to GAtt.}. Figure \ref{visual_work} visualizes the attention weights. Both RNNSearch and GAtt have very intuitive attentions, e.g. ``{\em refugees}'' is aligned to ``\chinese{难民}'', ``{\em government}'' is aligned to ``\chinese{政府}''. However, compared against those of RNNSearch, the attentions learned by GAtt are more focused and accurate. In other words, the refined source representations in GAtt help the attention mechanism concentrate its weights on translation-related words. 

To verify this point, we evaluated the quality of word alignments induced from different neural systems in terms of alignment error rate (AER)~\cite{Och:2003:SCV:778822.778824} and the soft version (SAER) of AER, following Tu et al.~\cite{DBLP:journals/corr/TuLLLL16}.\footnote{Notice that we used the same dataset and evaluation script as Tu et al.~\cite{DBLP:journals/corr/TuLLLL16}. We refer the readers to \cite{DBLP:journals/corr/TuLLLL16} for more details.} Table \ref{align_quality} display the evaluation results of word alignments. We find that both GAtt and GAtt-Inv significantly outperform RNNSearch in terms of both AER and SAER. Specifically, GAtt obtains a gain of 7.91 SAER and 7.3 AER points over RNNSearch. As we obtain word alignments by connecting target words to source words with the highest alignment probabilities computed according to their attention weights, the consistent improvements of our model over RNNSearch on AER score indicate that our model indeed learns more accurate attentions.

Another very important question is whether GAtt enhances the discrimination of the context vectors. We answer this question by visualizing these vectors, as shown in Figure \ref{context_visual}. We can observe that the heatmap of RNNSearch is very smooth, which varies very slightly across different decoding steps (the horizontal axis). This means that these context vectors are very similar to one another, thus lacking of discrimination. In contrast, there are obvious variations in GAtt. Statistically, the mean variance of the context vectors across different dimensions in RNNSearch is 0.0057, while it is 0.0365 in GAtt, 6 times larger than that of RNNSearch. Additionally, across different decoding steps, the mean variance is 0.0088 in RNNSearch, while it is 0.0465 in GAtt. All these strongly suggest that our model makes the context vectors more discriminative across different target words.

\subsection{Over-Translation Evaluation}\label{over_translation_analysis_sec}

Over-translation or repeatedly predicting the same target words~\cite{DBLP:journals/corr/TuLLLL16} is a challenging problem for NMT. We conjecture that the reason behind the over-translation issue is partially due to the small differences in context vectors learned by the vanilla attention mechanism. As the proposed GAtt can improve the discrimination power of context vectors, we hypothesize that our model  can deal better with the over-translation issue than the vanilla attention network. To testify this hypothesis, we introduce a metric called {\em N-Gram Repetition Rate} (N-GRR) that calculates the portion of repeated n-grams in a sentence:
\begin{equation}
\text{N-GRR} = \frac{1}{CR}\sum_{c=1}^{C}\sum_{r=1}^{R}\frac{|\text{N-grams}_{c,r}| - |u(\text{N-grams}_{c,r})|}{|\text{N-grams}_{c,r}|}
\end{equation}
where $|\text{N-grams}_{c,r}|$ denotes the number of total n-grams in the $r$-th translation of the $c$-th sentence in the testing corpus and $u(\text{N-grams}_{c,r})$ the number of n-grams after duplicate n-grams are removed. In our test sets, there are $C=6606$ sentences with $r=4$ and $r=1$ translations for the {\em Reference} and {\em NMT systems} respectively. If we compare N-GRR scores of machine-generated translation against those of reference translations, we can roughly know how serious the over-translation problem is. 

We show N-GRR results in Table \ref{over_translation_analysis}. Compared with reference translations (Reference), RNNSearch yields significant high scores, indicating that RNNSearch generates redundant repeated n-grams in translations, and therefore the over-translation problem in RNNSearch is serious. In contrast, both GAtt and GAtt-Inv achieve considerable improvements over RNNSearch in terms of N-GRR. Especially, we find that GAtt-Inv performs better than GAtt on all n-grams, which is in accordance with the translation results in Table \ref{test_performance}. These N-GRR results strongly suggest that the proposed models are able to handle the over-translation issue and that generating more discriminative context vectors makes NMT suffer less from the over-translation issue.
\begin{table}[t]
\begin{center}
{\small
\begin{tabular}{c||c|c|c|c}
{\bf System} & {\bf 1-gram} & {\bf 2-gram} & {\bf 3-gram} & {\bf 4-gram} \\
\hline
\hline
{\em Reference} & 12.94 & 1.80 & 0.93 & 1.29 \\
{\em RNNSearch} & 19.12 & 5.26 & 3.27 & 2.97 \\
\hline
{\em GAtt} & 18.09 & 4.11 & 2.50 & 2.46 \\
{\em GAtt-Inv} & 16.79 & 3.39 & 1.99 & 1.94 \\
\end{tabular}
}
\end{center}
\caption{\label{over_translation_analysis} N-GRR scores of different systems on all test sets with N ranges from 1 to 4. The lower the score, the better the system deals with the over-translation problem.}
\end{table}
\section{Conclusion}

In this paper, we have presented a novel GRU-gated attention model (GAtt) for NMT. Instead of using decoding-invariant source representations, GAtt produces new source representations that vary across different decoding steps according to the partial translation so as to improve the discrimination of context vectors for translation. This is achieved by a gating layer that regards the source representations and previous decoder state as the history and input to a gated recurrent unit. Experiments on Chinese-English translation tasks demonstrate the effectiveness of our model. In-depth analysis further reveals that our model is able to significantly reduce repeated redundant translations (over-translations).

In the future, we would like to apply our model to other sequence learning tasks as our model is easily to be adapted to any other sequence-to-sequence tasks (e.g. document summarization, neural conversion model, speech recognition, .etc). Additionally, except for the GRU unit, we will explore more different end-to-end neural architectures, such as convolutional neural network, LSTM unit as the gate mechanism plays a very important role in our model. Finally, we are interested in adapting our GAtt model as a tree-structured unit to compose different nodes in a dependency tree.

\newpage

\bibliographystyle{named}
\bibliography{acl2017}

\end{document}